\documentclass[sigconf, nonacm]{acmart}
\AtBeginDocument{%
  }

\usepackage{graphicx}
\usepackage{subcaption}   
\usepackage{booktabs}     

\DeclareMathOperator*{\argmax}{arg\,max}

\begin{document}

\title{Gaussian Process Bandits for Constraint Black Box Lever Optimization}

\author{Bram van den Akker}
\email{bram.vandenakker@booking.com}
\affiliation{%
  \institution{Booking.com}
  \city{Amsterdam}
  \country{The Netherlands}
}

\author{Hidde Fokkema}
\email{hidde.fokkema@gmail.com}
\authornote{Work done during an internship at Booking.com.}
\affiliation{%
  \institution{University of Amsterdam}
  \city{Amsterdam}
  \country{The Netherlands}
}

\author{Anastasiia Kornilova}
\email{
anastasiia.kornilova@booking.com}
\affiliation{%
  \institution{Booking.com}
  \city{Amsterdam}
  \country{The Netherlands}
}

\renewcommand{\shortauthors}{van den Akker, Fokkema \& Kornilova}

\begin{abstract}
Optimizing discrete allocation weights across competing segments of a shared budget is a fundamental challenge in resource-constrained, sequential decision-making environments. In practice, this optimization is often hindered by delayed, noisy, and aggregate feedback, alongside strict operational constraints on how frequently weights can be adjusted. Because of the resulting combinatorial action space and extreme data sparsity, classical multi-armed bandit algorithms that evaluate arms independently are rendered highly inefficient. To overcome these limitations, we propose a segmentation-agnostic approach utilizing Gaussian Process Upper Confidence Bound (GP-UCB) algorithms. By placing a Gaussian Process prior on the reward function, our method effectively shares information across neighboring configurations, drastically reducing the necessary exploration horizon. In this paper, we bridge the gap between theoretical bandit research and production-scale deployment. We first introduce a high-fidelity synthetic simulator to evaluate various kernel configurations under stationary, drifting, and reactive environments, identifying a composite Matérn + Linear kernel as the most robust choice for overcoming local noise and global shifts. Finally, we validate our findings through a large-scale Randomized Controlled Trial (RCT) in a live production environment. Our empirical results demonstrate that the GP-UCB approach delivers a statistically significant efficiency lift compared to a standard $\varepsilon$-greedy baseline, offering a highly practical solution for real-world black-box lever optimization.
\end{abstract}

\maketitle

\section{Introduction}

Consider a decision-maker who must periodically set discrete allocation weights across competing segments of a shared budget, then wait days or weeks for a single noisy aggregate outcome before being allowed to adjust the weights again. With even a handful of segments and modest discretization, the combinatorial action space can explode to thousands of configurations, yet we observe only tens of aggregate outcomes over the entire optimization horizon. This extreme mismatch between action-space size and data availability renders classical bandit algorithms, which require repeated per-arm sampling, fundamentally unsuitable.

This problem structure arises across diverse operational domains: in \emph{marketing mix optimization}, firms allocate spend across channels and observe delayed conversion signals~\citep{naik2003understanding}; in \emph{automated bidding for digital advertising}, platforms expose segment-level multipliers that steer budget towards audience subgroups~\citep{aggarwal2019autobidding, deng2021towards}; and in \emph{adaptive clinical trials}, limited enrollment slots are distributed across treatment arms based on interim outcomes~\citep{villar2015multi}; 

Despite the diversity of these applications, they share a common structure: segments draw from a unified budget, so an incorrect weight in one segment directly degrades the performance of others. The optimization problem is therefore coupled and combinatorial. Furthermore, practical constraints restrict how rapidly weights may change and limit feedback to aggregate, periodic observations and thus creating a high-noise, low-frequency decision environment.

While Multi-Armed Bandits~\citep{lattimore2020bandit} (MAB) and Reinforcement Learning~\citep{sutton1998reinforcement} (RL) are common approaches for sequential decision-making, they face significant practical challenges in such settings. Decision-makers operate under strict data constraints, receiving only aggregated, delayed feedback. Furthermore, operational policies require adjustments to be gradual and infrequent (typically on a weekly or bi-weekly cadence).
To navigate this environment, we propose a segmentation-agnostic method based on Gaussian Process Multi-Armed Bandits~\citep{srinivas2010gaussian, krause2011contextual}.

In this paper, we demonstrate the practical application of Gaussian Process Upper Confidence Bound (GP-UCB)~\citep{srinivas2010gaussian} algorithms to optimize budget allocation under the strict constraints of real-world operating environments. We bridge the gap between theoretical bandit research and production-scale deployment by providing a simulation-to-deployment pipeline.
Our key contributions are as follows:
\begin{itemize}
\item \textbf{Environment Simulation:} We develop a high-fidelity synthetic simulator that models the key complexities of black-box allocation systems, including observation noise, reward drift, and reactive (adversarial) dynamics. This environment serves as a benchmark for testing agent robustness before live deployment.
\item \textbf{Offline-to-Online Evaluation Framework:} We propose a methodology for kernel selection and hyperparameter optimization using synthetic regret curves.
\item \textbf{Live Production Validation (RCT):} We report results from a large-scale Randomized Controlled Trial (RCT) in a live production environment. Our empirical results confirm the model’s efficacy, delivering a statistically significant performance lift in the primary efficiency metric relative to the standard $\varepsilon$-greedy baseline.
\end{itemize}

\section{Related Work}

\paragraph{Gaussian Process bandits.} 
The GP-UCP algorithm, introduced by Srinivas et al.\ \citep{srinivas2010gaussian},
extends the classical UCB algorithm for finite
arms \citep{auer2002finite} to continuous domains by leveraging the mean and
covariance function of the Gaussian Process posterior. The GP model captures
dependencies between arms that are close in the action space, and its posterior
variance provides calibrated uncertainty estimates, together yielding a
principled exploration–exploitation trade-off over the continuous action set.
Following its
introduction, the GP-UCB algorithm was further improved and extended to
improve the theoretical regret bounds, incorporate contextual information and
dynamic environments \citep{chowdhury2017kernelized, krause2021context,
pmlr-v51-bogunovic16}. GP-based optimization algorithms have recently been
applied to online marketing problems such as bid optimization \citep{castiglioni2025bid} and
budget allocation \citep{hu2025optimal}.  

\paragraph{Bandit convex optimization.} 
Bandit optimization over continuous rewards has a rich history of theoretical research 
going back to the work of Flaxman, Kalai
and McMahan \citep{flaxman2005gradient}. In particular, when assuming that the
reward function is convex, there is an extensive stream of work on gradually
improving the theoretical upper bound and computational complexity,
\citep{hazan2014bandit, bubeck2021kernel, fokkema2024online,
pmlr-v247-suggala24a}. A comprehensive overview of these results can be found
in \citep{Lattimore_2026}. When the number of data samples becomes large,
recent algorithms for bandit convex optimization would be computationally
appealing over GP-UCP. We explored the usage of Convex Bandits in our preliminary study. However, we were unable to find evidence that our reward landscapes are convex and that we did not have sufficient samples to obtain an edge in the computation. Moreover, we did not observe competitive results, and therefore we do not further discuss Bandit convex optimization in our work.
\section{Methodology}\label{sec:methodology}
Let $\lambda \in [0, 1]^d$ represent the vector of allocation weights for $d$ distinct segments. To simplify our search space, we constrain these weights   to discretized values within a fixed range (e.g., $[0.10, 1.0]$) using fixed increments (e.g., $0.05$). 
Due to operational limitations, these weights can only be adjusted periodically (e.g., once per week), after which we observe a single aggregated reward metric.
Optionally, we can subject the weights to a step-size limit $\Delta_{\max}$.

Our objective is to maximize the efficiency of budget utilization across targeted segments. At each time step $t$, we select an action $A_t \in [0.10, 1.0]^d$ and observe a noisy reward $Y_t$, formalized as:
\begin{equation}
    Y_t = f(A_t) + \varepsilon_t
\end{equation}
where $f$ represents the true, unobserved reward landscape and $\varepsilon_t$ denotes observational noise.

The main challenge of this setup is the extreme data sparsity. We may receive only a single aggregated reward observation (summed across all segments) per week, making it impossible to attribute performance to individual segment weights. Furthermore, even when the continuous treatments are simplified to discrete actions in $0.05$ increments, this creates a vast combinatorial action space. For example, an action space between $0.10$ and $1.0$ with $0.05$ increments, and just 3 segments results into $6,859$ unique actions, taking $131$ years to try each action at a once-per-week cadence.

This implies that if we treat each weight configuration (arm) as strictly independent, the problem becomes unsolvable within a realistic time horizon. Standard multi-armed bandit (MAB) algorithms, such as Upper Confidence Bound (UCB) and Thompson Sampling (TS), rely heavily on repeated sampling to build confidence bounds for individual arms. Consequently, UCB and TS algorithms are forced into a state of near-perpetual exploration, performing substantially worse in terms of cumulative regret than simple $\epsilon$-greedy strategies.

To overcome the limitations of independent arm evaluation, we must exploit the underlying structure of the reward function. It is reasonable to assume that the reward landscape is continuous and that similar multiplier allocations yield similar returns.

\subsection{Gaussian Process Bandits}
One method of exploiting this structural dependency is by using Gaussian Process (GP) Bandits. In this setup, we place a Gaussian Process prior on the reward function, $f \sim \mathcal{GP}(\mu, k)$, where $\mu$ is the mean function and $k$ is the covariance kernel. This allows the algorithm to infer the expected reward of unexplored multiplier configurations by leveraging the observed rewards of neighboring actions, effectively sharing context across the domain. The most common approach for this is the GP Upper Confidence Bound (GP-UCB)~\citep{srinivas2010gaussian} algorithm. At each step, the GP-UCB algorithm balances exploitation and exploration by selecting the arm that maximizes the upper confidence bound of the predictive posterior distribution:
\begin{align}\label{eq:arm-choice}
    A_t = \argmax_{a \in [0.10, 1.0]^{d}}\mu_{t -1}(a) + \beta_t \sigma_{t - 1}(a) 
.\end{align}
Here, $\mu_{t-1}$ and $\sigma_{t-1}$ are the posterior mean and standard
deviation of the posterior of all data up to previous round. The parameter
$\beta_t$ is an exploration parameter, which ensures that enough exploration 
occurs even after many steps.

\subsubsection{Kernel Selection}
The efficiency of the GP-UCB algorithm is highly dependent on the choice of the kernel function $k$, which encodes our prior assumptions about the smoothness and shape of $f$ \citep{kanagawa2018gaussian}. We initially consider several independent kernel choices: 

\begin{itemize}
    \item \textbf{Radial Basis Function (RBF)~\citep{rasmussen2006gaussian}:} Utilized for smooth, infinitely differentiable reward approximations.
    \item \textbf{Matérn Kernel~\citep{matern1960spatial}:} Evaluated with varying $\nu$ parameters to accommodate rougher, less predictable underlying reward landscapes.
    \item \textbf{Linear Kernel:} Employed independently to model straightforward, monotonic trends across the reward space.
\end{itemize}

Additionally, the mechanics of the platform suggest inherent monotonicity; increasing the budget allocation for a segment generally reduces its marginal ROI due to audience saturation and diminishing returns from reaching less responsive user segments. To explicitly enforce this structural assumption while maintaining the expressive power of non-linear models, we evaluate composite kernels by adding a linear component to the base kernels (i.e., $k_{\text{RBF}} + k_{\text{Linear}}$ and $k_{\text{Matérn}} + k_{\text{Linear}}$). This additive structure effectively regularizes the predicted landscape and improves algorithmic convergence. The algorithm's behavior is further tuned via the \textit{length scale} parameter, which dictates spatial correlation.

\section{Experimental Setup}\label{sec:experimental-setup}

We evaluate the proposed GP-UCB framework in a simulator that
aims to capture the key dynamics of black-box budget allocation
optimization. The simulator lets us (i) control the reward landscape,
(ii) introduce non-stationarity and adversarial effects, and (iii)
measure regret against an oracle. This level of counterfactual analysis is impossible with real-world A/B tests alone.

\subsection{Simulated Allocation Environment}\label{sec:sim-env}

\paragraph{Proportional allocation.}
We model a decision-maker that distributes a fixed periodic budget~$X$ across
$d$~segments.  Each segment~$i$ is assigned an \emph{allocation
weight}~$\lambda_i$, and the system allocates spend
proportionally:
\begin{equation}\label{eq:allocation}
  X_i \;=\; X \,\cdot\, \frac{\lambda_i}{\sum_{k=1}^{d} \lambda_k}\,.
\end{equation}
This represents a common allocation mechanism where weights steer
resources towards segments with higher expected returns.

\paragraph{Reward landscape.}
For each segment~$i$ we construct a smooth, non-monotone reward curve
$r_i(\lambda_i)$ using a sum of Gaussian \emph{inflection points}.
Concretely, let $\{(x_j, s_j, w_j)\}_{j=1}^{J_i}$ be the set of
inflection-point parameters (location, steepness, width) for
segment~$i$.  We first form the derivative
\begin{equation}\label{eq:derivative}
  r_i'(\lambda) \;=\;
    \sum_{j=1}^{J_i} -\,s_j \,\exp\!\Bigl(
      -\frac{(\lambda - x_j)^2}{2\,w_j^2}
    \Bigr),
\end{equation}
then obtain $r_i$ by numerical integration (cumulative trapezoidal
rule) and min-max scaling to a target range $[y_{\min},\, y_{\max}]$.
The number of inflection points per segment is sampled uniformly from
$\{5,\dots,9\}$; steepness values are drawn from $\mathcal{U}(0,1)$
and widths from $\mathcal{U}(0, 0.1)$.

\begin{figure}[htbp]
    \centering
    \includegraphics[width=\columnwidth]{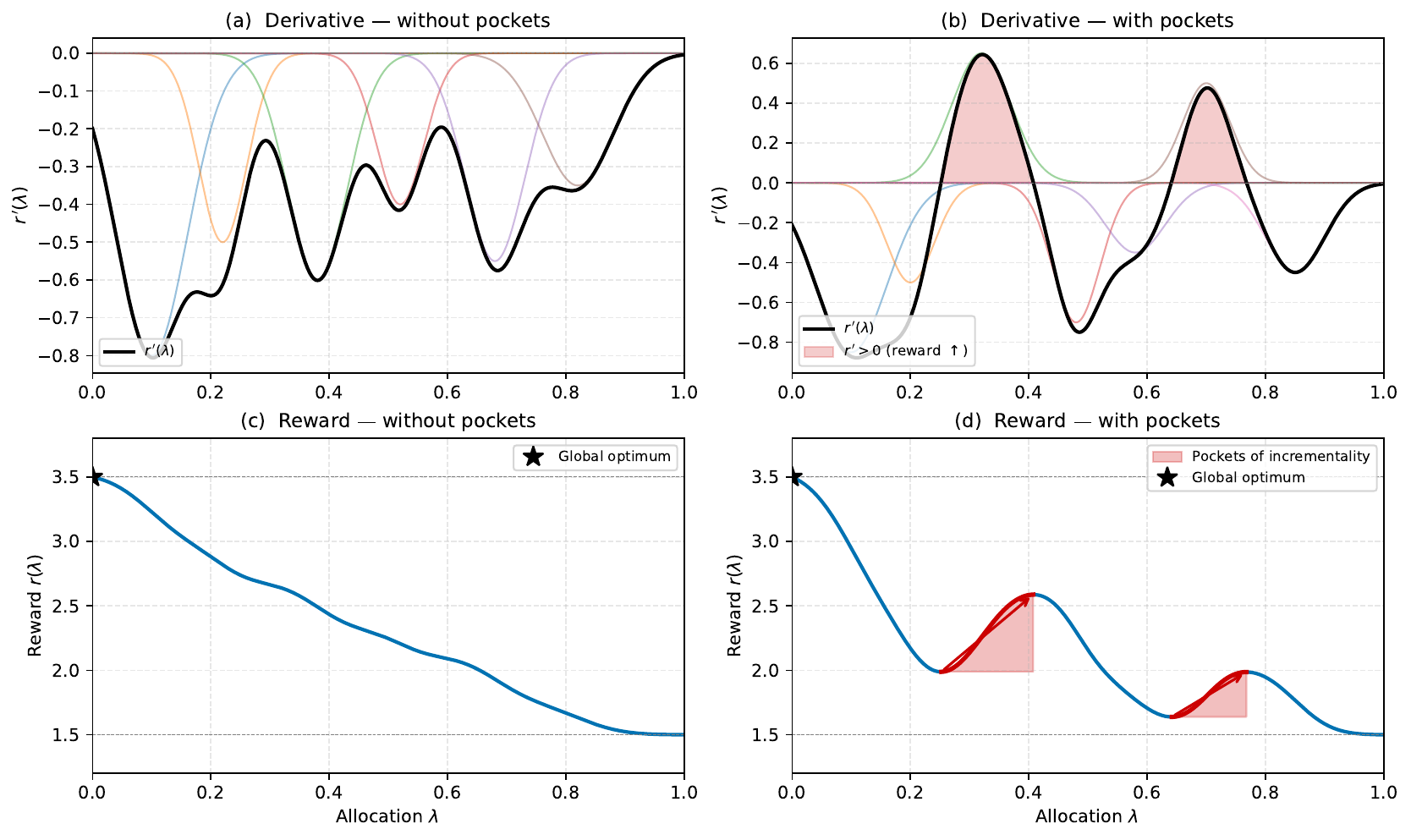}
    \caption{Reward landscapes are constructed using gaussian \textit{inflection points} as shown (a) and (b). These inflection points represent the derivatives of the reward landscape and result into a non-linear function, which could be either monotonically increasing/decreasing (c) or cause \textit{pockets of incrementality} where additional allocation results in incremental rewards. }
    \label{fig:reward_landscape}
\end{figure}

This construction produces landscapes with a variety of slopes, including occasional regions of diminishing returns, reflecting the empirical observation that black-box allocation systems can leave exploitable
\emph{pockets of incrementality} as shown in Figure~\ref{fig:reward_landscape}.

\begin{figure*}[t]
    \centering
    \includegraphics[width=\textwidth]{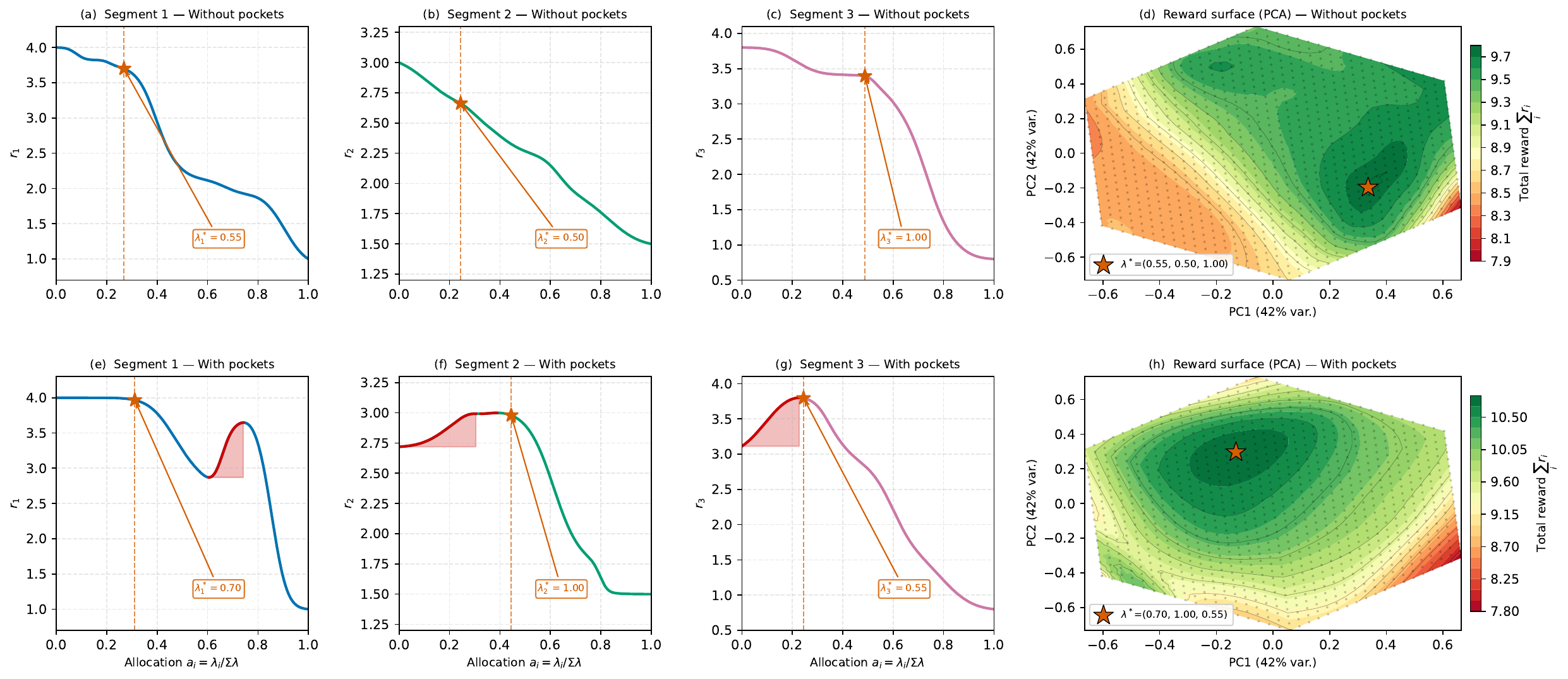}
    \caption{When segments share a budget (proportional allocation), the combined reward surface can have local optima even if individual segment curves are monotonic, because shifting weight between segments creates interaction effects. The top row shows a set of segment without \textit{pockets of incrementality}, and the bottom row shows landscapes with multiple segments. We have marked the optimal points for each segment and their combined landscape with a star. Our goal is to find these points.}\label{fig:reward_landscape_simulation}
\end{figure*}

In Figure~\ref{fig:reward_landscape_simulation}, we highlight how multiple segments collectively form a ground truth reward landscape with multiple local optima. Even if there are no pockets of incrementality in the individual segments, the reallocation of budget can result into increased efficiency of the overall return on investment.

\paragraph{Action space and constraints.}
Multipliers are discretised to $\lambda_i \in
\{0.10,\, 0.15,\, \dots,\, 1.00\}$ (19~values per segment).  Two
operational constraints restrict the agent at each step:
\begin{enumerate}
  \item \textbf{Step-size limit.}
    $\Delta_{\max} = 0.10$ (at most two discrete increments per segment per step), preventing drastic week-to-week changes that could destabilise system performance.
  \item \textbf{Anchor constraint.}
    At least one segment must remain at $\lambda_i = 1.0$. This prevents configurations that are scalar multiples of each other from appearing as distinct arms (e.g., $[0.1, 0.1, 0.1]$ and $[0.2, 0.2, 0.2]$ yield identical allocations under Equation~\ref{eq:allocation}).
\end{enumerate}
These constraints mirror real-world operational policies and reduce the
effective action space considerably.

\paragraph{Noise.}
Observed rewards are corrupted by additive Gaussian noise,
$\tilde{r}_i = r_i + \varepsilon$ with
$\varepsilon \sim \mathcal{N}(0, \sigma^2)$.  The noise scale~$\sigma$
is sampled uniformly from $[0.2, 0.5]$ per experiment, simulating the
inherent measurement uncertainty in weekly outcome metrics.

\subsection{Environment Variants}\label{sec:env-variants}

A key question in practice is whether a kernel that performs well under
stationary conditions retains its advantage when the reward landscape
shifts over time, either organically or adversarially.  We therefore
evaluate four environment types of increasing difficulty.

\paragraph{Base (noiseless)}
The reward landscape is fixed and observations are noise-free
($\sigma = 0$).  This idealised setting isolates the pure benefit of
GP-based exploration over a stationary, deterministic reward surface
and serves as a best-case reference.

\paragraph{Stationary with noise (NOISE)}
The reward landscape is fixed for the entire horizon; the only source
of uncertainty is observation noise.  This serves as the base case.

\paragraph{Non-stationary (DRIFT)}
Each inflection point undergoes a random walk:
every $T_{\mathrm{shift}}$ steps, its parameters are perturbed as
\begin{equation}\label{eq:drift}
  x_j \leftarrow x_j + \eta_x,\qquad
  s_j \leftarrow s_j + \eta_s,\qquad
  w_j \leftarrow w_j + \eta_w,
\end{equation}
where $\eta_\cdot \sim \mathcal{N}(0, \sigma_{\mathrm{shift}}^2)$.  We
set $T_{\mathrm{shift}} = 10$~steps and
$\sigma_{\mathrm{shift}} = 0.1$, yielding gradual landscape drift that
invalidates earlier observations.

\paragraph{Reactive / adversarial (REACTIVE)}
In practice, allocation environments are rarely fully stationary: competing advertisers may react to observed market shifts, and platform dynamics can create feedback loops that partially penalize predictable policies. Rather than modelling a fully adversarial game (which would require minimax or game-theoretic methods), we introduce a mildly reactive environment as a robustness stress-test. The goal is not to find the optimal policy under adversarial conditions, but to assess whether a kernel's advantage degrades gracefully when the environment partially counteracts the agent's strategy.  At each
step the environment:
\begin{enumerate}
  \item pre-samples $K{=}1{,}000$ candidate adversarial landscapes;
  \item predicts the agent's next action $\hat{\boldsymbol{\lambda}}_{t+1}$
        (using the agent's most recent choice as a proxy);
  \item selects the candidate whose worst-performing point is closest to
        $\hat{\boldsymbol{\lambda}}_{t+1}$;
  \item blends the selected adversarial landscape with the base landscape
        via a mixture parameter $\alpha$:
\end{enumerate}
\begin{equation}\label{eq:reactive}
  r_i^{(\mathrm{eff})}(\lambda)
  \;=\;
  (1-\alpha)\,r_i^{(\mathrm{base})}(\lambda)
  \;+\;
  \alpha\,r_i^{(\mathrm{adv})}(\lambda),
\end{equation}
The mixture parameter $\alpha = 0.2$ ensures that 80\% of the base landscape is preserved, modelling a setting where competitive reactions (e.g.\ shifting competitor pricing or competing agents) create headwinds but do not fully dictate the reward. This represents a realistic middle ground: environments are rarely purely stationary, but also rarely fully adversarial.

\subsection{Algorithms}\label{sec:algorithms}

\paragraph{GP-UCB variants.}
We instantiate the GP-UCB algorithm (Section~\ref{sec:methodology})
with five kernel configurations:
\begin{itemize}
  \item \textbf{RBF}\,: squared-exponential kernel with length scale
        $\ell = 1.5$;
  \item \textbf{Mat\'ern}\,: Mat\'ern kernel with $\ell = 1.2$,
        $\nu = 1.5$;
  \item \textbf{Linear}\,: dot-product kernel
        ($\sigma_0 = 0$);
  \item \textbf{RBF\,+\,Linear}\,: sum of RBF ($\ell{=}1.5$) and
        dot-product ($\sigma_0{=}1$);
  \item \textbf{Mat\'ern\,+\,Linear}\,: sum of Mat\'ern
        ($\ell{=}1.2$, $\nu{=}1.5$) and dot-product
        ($\sigma_0{=}1$).
\end{itemize}
Composite (sum) kernels allow the GP to capture both smooth local
structure \emph{and} a global linear trend simultaneously.  All
variants use $\delta = 0.1$ in the exploration parameter
$\beta_t = 1 + \sqrt{2(\gamma_t + 1 + \ln(1/\delta))}$, with
$\gamma_t = \tfrac{1}{2}(1 + \ln t)^2$ as an upper bound on the
maximum information gain.

\paragraph{$\varepsilon$-greedy baseline.}
We compare against an $\varepsilon$-greedy policy with a decreasing
exploration rate:
\begin{equation}\label{eq:epsilon}
  \varepsilon_t \;=\; \min\!\bigl(1,\;
    t^{-1/3}\,(\ln t)^{1/3}\bigr),
\end{equation}
which ensures sub-linear regret in multi-armed bandit settings.  At each step the agent either explores
uniformly at random or exploits the arm with the highest observed mean
reward.

\paragraph{Reference baselines.}
Two additional baselines bracket performance:
\begin{itemize}
  \item \textbf{Constant}: holds $\boldsymbol{\lambda}_t =
        (1,1,\dots,1)$ throughout, representing a decision-maker that does
        not optimize weights at all.
  \item \textbf{Optimal}: an oracle that knows the true reward landscape
        and follows the shortest path to the global optimum
        under the step-size constraint (stationary setting) or
        greedily tracks the per-step optimum (non-stationary and
        reactive settings).
\end{itemize}

\paragraph{Why UCB and Thompson Sampling are omitted.}
Classical multi-armed bandit algorithms such as UCB and Thompson
Sampling treat each arm independently.  In our setting, the
combinatorial action space contains $|\mathcal{A}| = 6{,}859$ feasible
arms (after applying the anchor constraint), yet the decision horizon
is only $T = 52$ weeks.  Standard bandit methods would therefore spend
the entire horizon exploring and thus incur near-maximal regret. GP-UCB circumvents this by
generalising across arms via the kernel, requiring far fewer
observations to identify high-reward regions.

\subsection{Evaluation Methodology}\label{sec:evaluation}

\paragraph{Regret.}
We measure performance via \emph{cumulative regret}:
\begin{equation}\label{eq:regret}
  R(T) \;=\; \sum_{t=1}^{T}
  \bigl[r^t(\boldsymbol{\lambda}_t^*) - r^t(\boldsymbol{\lambda}_t)\bigr],
\end{equation}
where $\boldsymbol{\lambda}_t^*$ is the (possibly time-varying) optimal arm of
the rewards and $r^t(\cdot)$ denotes the expected (noise-free) reward at round
$t$.  To highlight relative improvement over a na\"ive strategy, we also report
\emph{difference regret}: $R_{\mathrm{model}}(T) -
R_{\mathrm{Constant}}(T)$, which is negative when a model outperforms
the Constant baseline.

\paragraph{Protocol.}
Each configuration is repeated over $N = 100$ independent experiments,
each with a freshly sampled reward landscape and noise realisation.
The horizon is $T = 52$~steps (one year of weekly decisions) with
$d = 3$~segments.  We report means and standard errors
($\text{std}/\!\sqrt{N}$) across experiments.  The focus on
non-stationarity with noise, adversarial, and reactive environments
allows us to assess the \emph{generalisability} of each kernel across
fundamentally different forms of distributional shift. To the best of our knowledge, this comparison has not been systematically
explored in the Gaussian process bandit optimization literature.




\section{Results}

\subsection{Offline Synthetic Results}\label{sec:results}

\begin{figure*}[t]
  \centering
  \begin{subfigure}[b]{0.48\linewidth}
    \includegraphics[width=\linewidth]{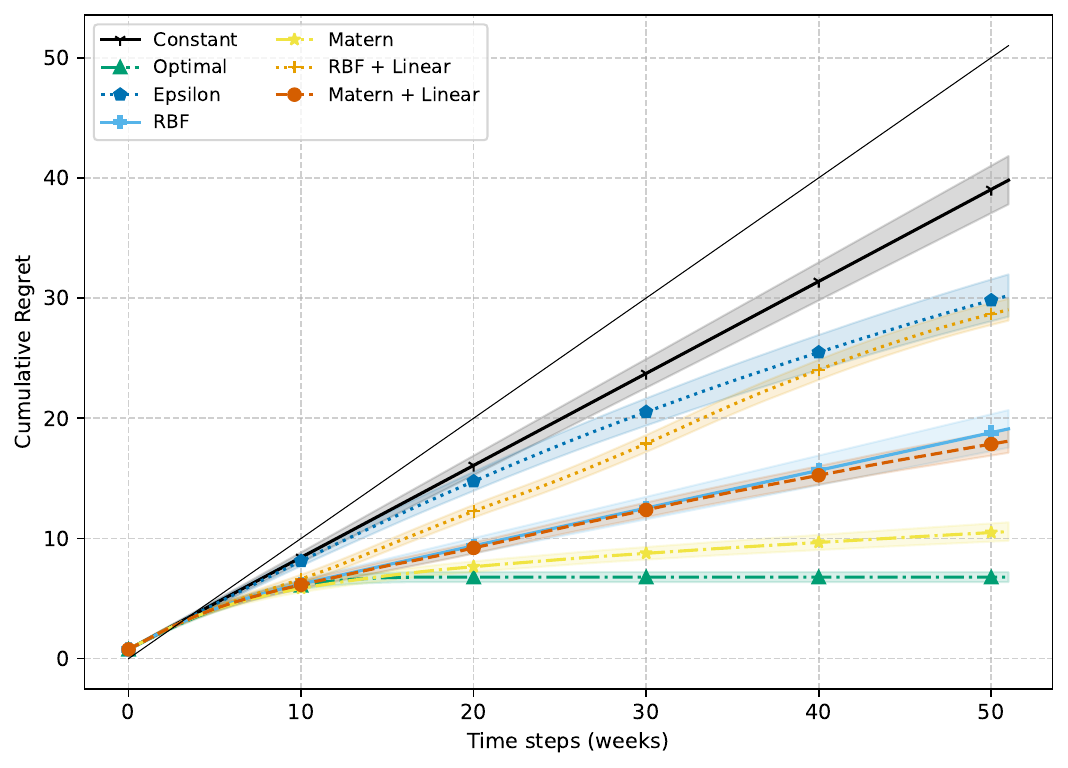}
    \caption{Base (noiseless)}
    \label{fig:regret-base}
  \end{subfigure}\hfill
  \begin{subfigure}[b]{0.48\linewidth}
    \includegraphics[width=\linewidth]{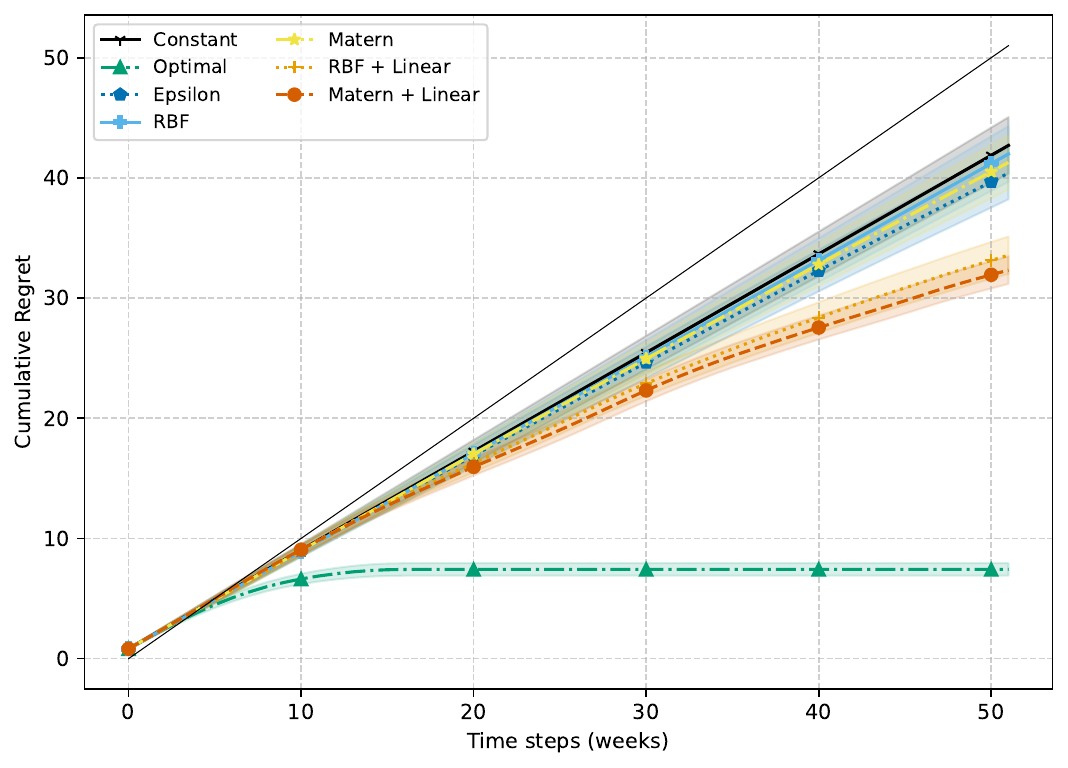}
    \caption{Stationary with noise}
    \label{fig:regret-noise}
  \end{subfigure}\\[4pt]
  \begin{subfigure}[b]{0.48\linewidth}
    \includegraphics[width=\linewidth]{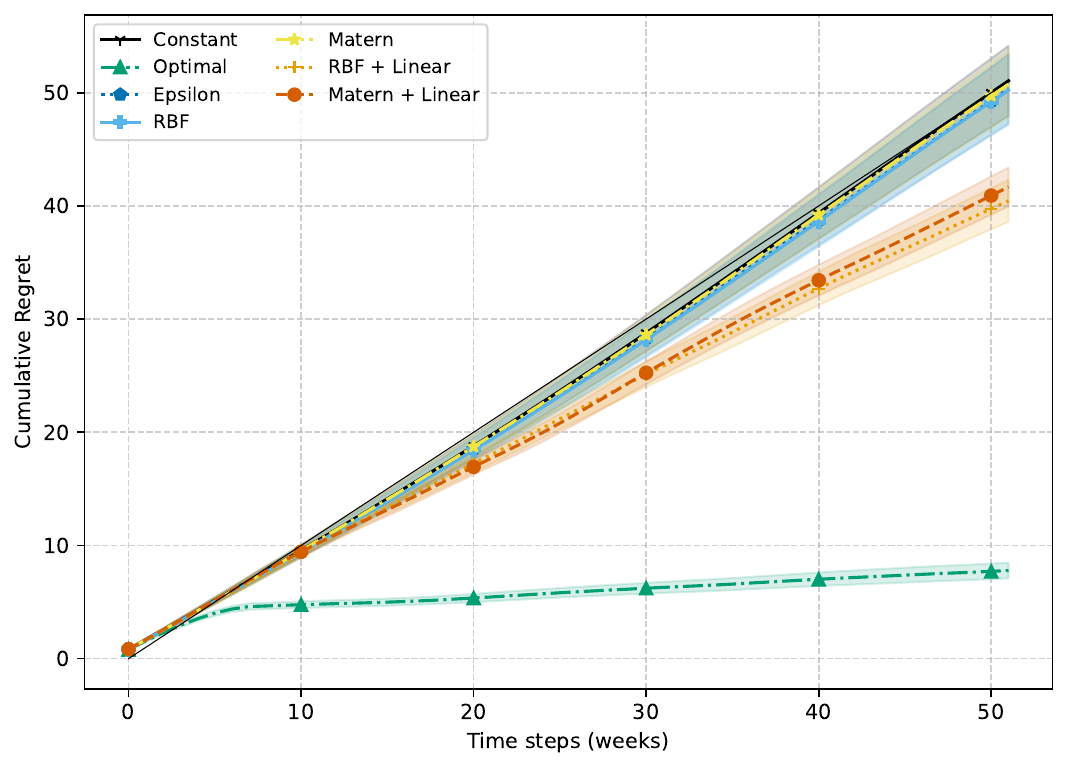}
    \caption{Non-stationary (drift)}
    \label{fig:regret-ns}
  \end{subfigure}\hfill
  \begin{subfigure}[b]{0.48\linewidth}
    \includegraphics[width=\linewidth]{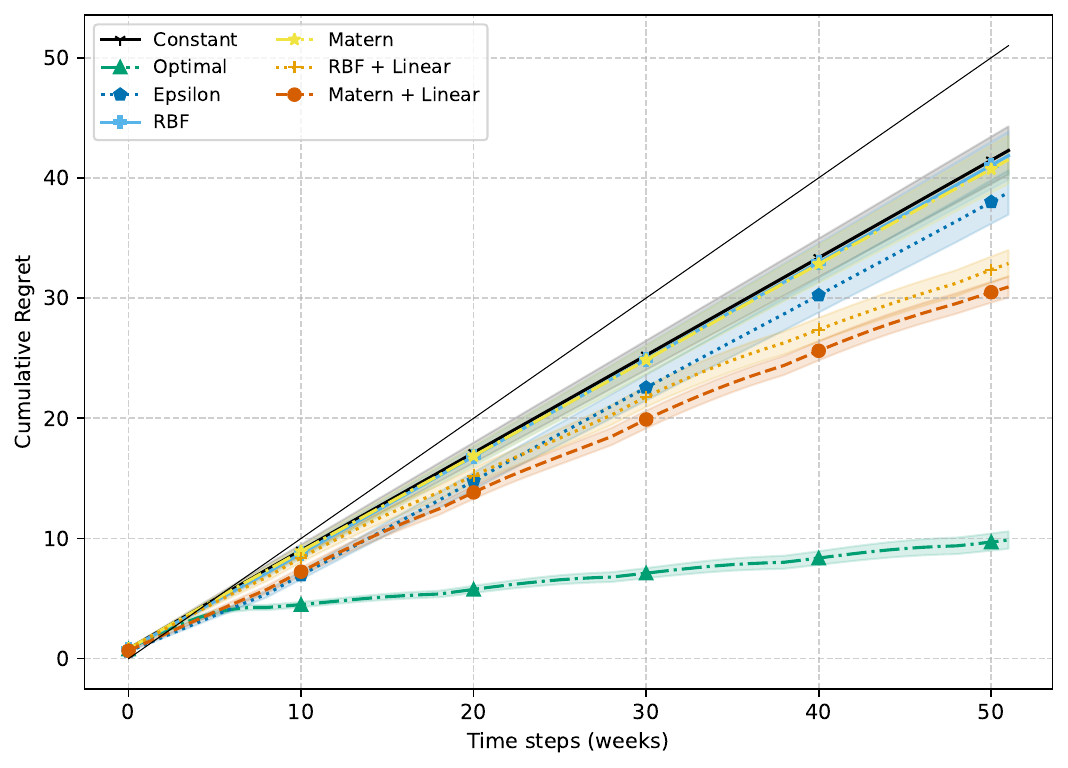}
    \caption{Reactive / adversarial}
    \label{fig:regret-reactive}
  \end{subfigure}
  \caption{Cumulative regret over $T{=}52$ weeks.  The black diagonal
    represents the regret when the reward is consistently zero.}
  \label{fig:cumulative-regret}
\end{figure*}

We compare all models from Section~\ref{sec:algorithms} across the
four environments of Section~\ref{sec:env-variants}, aggregating
$N{=}100$ experiments per configuration ($T{=}52$ steps each).
Shaded bands denote $\pm 1$~SEM.  Significance is assessed via
one-sided paired $t$-tests against $\varepsilon$-greedy.
Figure~\ref{fig:cumulative-regret} shows cumulative regret curves;
Table~\ref{tab:final-regret} summarises final regret and the first
step at which each GP-UCB variant significantly outperforms
$\varepsilon$-greedy.

\paragraph{Pure kernels are brittle.}
In the noiseless base environment, pure Mat\'ern achieves the lowest
regret of all models ($10.6$, a $65\%$ reduction vs.\
$\varepsilon$-greedy; significant from step~4), followed by RBF
($19.1$, $37\%$).  However, this advantage vanishes entirely once
any form of environment complexity is introduced.  Under observation
noise, drift, and adversarial reactions alike, pure RBF and Mat\'ern
regrets are statistically indistinguishable from
$\varepsilon$-greedy. However, under mild adversarial pressure, pure kernels are significantly worse than $\varepsilon$-greedy (p < 0.001), as the reactive environment exploits the GP's tendency to concentrate predictions in a narrow region. Composite kernels, by contrast, maintain their advantage, confirming their robustness beyond stationary conditions.

\paragraph{Composite kernels are robust.}
Augmenting either kernel with a linear component restores a
significant advantage across all four environments.  The linear term
captures a coarse global trend in the weight space, anchoring
exploration even when local GP estimates are degraded by noise or
distributional shift.  Both RBF\,+\,Linear and Mat\'ern\,+\,Linear
achieve $15$--$20\%$ regret reductions over $\varepsilon$-greedy in
every challenging setting (all $p < 10^{-3}$ at $T{=}52$).

\paragraph{Mat\'ern\,+\,Linear is the most consistent performer.}
Table~\ref{tab:final-regret} shows that Mat\'ern\,+\,Linear achieves
significance earliest across environments (steps~4, 24, 19, 23) and
delivers the largest or near-largest improvement in every setting.
RBF\,+\,Linear, by contrast, requires up to step~38 under
adversarial conditions and loses significance in the base case by
$T{=}52$.  The finite smoothness of the Matérn component ($\nu{=}1.5$) avoids the oversmoothing of RBF, while the linear component provides a stable global trend estimate that regularizes predictions under noisy observations. Together, the composite kernel balances local adaptability with global structure. Based on these
results, we select \textbf{Mat\'ern\,+\,Linear} as the candidate
kernel for the real-world experiment in Section~\ref{sec:online}.

\begin{table}[t]
\centering
\caption{Final cumulative regret at $T{=}52$ (mean $\pm$ SEM) and
  first step with significant improvement over $\varepsilon$-greedy
  ($p < 0.05$, one-sided paired $t$-test).
  ``---'': not significant; ``$\downarrow$'': significantly worse.}
\label{tab:final-regret}
\small
\setlength{\tabcolsep}{4pt}
\begin{tabular}{@{}l c c c c@{}}
\toprule
 & \textbf{BASE} & \textbf{NOISE} & \textbf{DRIFT} & \textbf{REACT.} \\
\midrule
Constant     & $39.8{\scriptstyle\pm2.0}$ & $42.7{\scriptstyle\pm2.4}$ & $51.1{\scriptstyle\pm3.1}$ & $42.3{\scriptstyle\pm2.0}$ \\
Optimal      & $\phantom{0}6.8{\scriptstyle\pm0.4}$  & $\phantom{0}7.4{\scriptstyle\pm0.5}$  & $\phantom{0}7.8{\scriptstyle\pm0.7}$  & $\phantom{0}9.9{\scriptstyle\pm0.7}$ \\
$\varepsilon$-greedy  & $30.2{\scriptstyle\pm1.8}$ & $40.4{\scriptstyle\pm2.2}$ & $50.3{\scriptstyle\pm3.1}$ & $38.8{\scriptstyle\pm1.9}$ \\
\midrule
RBF          & $19.1{\scriptstyle\pm1.6}$ & $42.0{\scriptstyle\pm2.3}$ & $50.2{\scriptstyle\pm3.1}$ & $41.9{\scriptstyle\pm2.0}$ \\
             & \textit{step\,4} & --- & --- & $\downarrow$ \\[1pt]
Mat\'ern     & $10.6{\scriptstyle\pm0.8}$ & $41.3{\scriptstyle\pm2.3}$ & $50.8{\scriptstyle\pm3.0}$ & $41.5{\scriptstyle\pm2.0}$ \\
             & \textit{step\,4} & --- & --- & $\downarrow$ \\[1pt]
RBF+Lin.     & $29.0{\scriptstyle\pm0.9}$ & $33.6{\scriptstyle\pm1.6}$ & $40.4{\scriptstyle\pm1.9}$ & $32.9{\scriptstyle\pm1.1}$ \\
             & \textit{step\,4}$^*$ & \textit{step\,31} & \textit{step\,20} & \textit{step\,38} \\[1pt]
Mat.+Lin.    & $18.1{\scriptstyle\pm1.0}$ & $32.3{\scriptstyle\pm1.2}$ & $41.7{\scriptstyle\pm1.7}$ & $30.9{\scriptstyle\pm0.9}$ \\
             & \textit{step\,4} & \textit{step\,24} & \textit{step\,19} & \textit{step\,23} \\
\bottomrule
\multicolumn{5}{@{}l}{\footnotesize ${}^*$Significant at step~4 but not at $T{=}52$ ($p = 0.24$).}
\end{tabular}
\end{table}



\subsection{Real world results}\label{sec:online}
Guided by the synthetic benchmarks in Section~\ref{sec:results}, we deployed the Mat\'ern\,+\,Linear GP-UCB model into a live production environment to validate its performance under true non-stationary conditions. Operational environments are characterized by heterogeneous trends and latent seasonality across individual segments; thus, a robust kernel is essential for stable exploration.
\paragraph{Experimental Design.} We implemented a Randomized Controlled Trial (RCT) with two variants:
\begin{itemize}
\item \textbf{Control:} Decisions governed by the $\varepsilon$-greedy multi-armed bandit (MAB) strategy described in Section~\ref{sec:algorithms}.
\item \textbf{Treatment:} Decisions governed by the GP-UCB model utilizing the Mat\'ern\,+\,Linear kernel.
\end{itemize}
The trial duration was set to $T{=}6$ weeks. This window was determined by a priori power calculations to ensure sufficient sensitivity to detect a meaningful performance lift, while also aligning with the synthetic results in Section~\ref{sec:results}. To improve discriminative power, models were initialized using parameters optimized from about 52 weeks of historical operational data. The newly collected data points were not shared among control and treatment.

\paragraph{Performance Lift.} The deployment results aligned with the synthetic findings\footnote{Due to commercial sensitivity, we report directional findings rather than absolute metrics. }, demonstrating GP-UCB’s ability to navigate complex reward landscapes more efficiently than the $\varepsilon$-greedy baseline. Our primary efficiency metric showed a substantial and statistically significant improvement. The treatment group exhibited a positive performance lift across all tested segments at the 90\% confidence level. Notably, the treatment group exhibited faster adaptation to short-term environmental fluctuations, likely facilitated by the Mat\'ern component’s ability to model local fluctuations without the rigid global smoothness of an RBF kernel.
\paragraph{Conclusion.}The real-world results validate the robustness of the composite Mat\'ern\,+\,Linear kernel. By anchoring local GP estimates with a global linear trend, the model maintained high performance despite the noise and drift inherent in live allocation environments. This confirms that the selection criteria established in Section~\ref{sec:results} translate effectively to improved operational outcomes.


\section{Conclusion \& future work}
This study addressed the challenge of optimizing allocation performance in highly non-stationary landscapes by bridging the gap between controlled simulation and live production. By developing a robust simulation environment that replicates the noise, drift, and adversarial dynamics of real-world allocation environments, we identified the Mat\'ern\,+\,Linear kernel as a superior alternative to traditional exploration strategies. Our simulation-first approach provided the necessary statistical confidence to deploy the model in a live Randomized Controlled Trial (RCT), where the GP-UCB framework delivered a statistically significant efficiency gain over the standard $\varepsilon$-greedy baseline.
\paragraph{Future Work.} Building on the current deployment's success, our simulation results reveal the potential for additional significant performance improvements. Preliminary results within the simulator indicate that explicit modeling of temporal non-stationarity and the inclusion of contextual features (e.g., segment characteristics, environment-specific signals) can further reduce regret. However, the transition from a Multi-Armed Bandit (MAB) to a Contextual Bandit framework increases the parameter space exponentially, necessitating further research in the following areas:
\begin{itemize}
\item \textbf{High-Dimensional Hyperparameter Optimization:} We intend to leverage our simulator to conduct large-scale sensitivity analyses, identifying which contextual variables provide the highest information gain relative to the increased computational and statistical complexity.
\item \textbf{Dynamic Contextual Modeling:} Future iterations will explore the use of spatio-temporal kernels designed to adapt to rapid distribution shifts, converting the promising results observed in our "Drift" and "Reactive" simulations into a scalable, real-world contextual experiment.
\item \textbf{Automated Kernel Selection:} Investigating automated methods to adaptively weight the linear and non-linear components of the kernel as the environment's volatility changes over time.
\end{itemize}


\bibliographystyle{ACM-Reference-Format}
\bibliography{main}

@book{rasmussen2006gaussian,
    title     = {Gaussian Processes for Machine Learning},
    author    = {Rasmussen, Carl Edward and Williams, Christopher K. I.},
    year      = {2006},
    publisher = {MIT Press},
    address   = {Cambridge, MA}
  }

@book{Lattimore_2026, 
    place={Cambridge},
    title={Bandit Convex Optimisation},
    publisher={Cambridge University Press}, 
    author={Lattimore, Tor}, 
    year={2026}
}

@article{kanagawa2018gaussian,
  title={Gaussian processes and kernel methods: A review on connections and equivalences},
  author={Kanagawa, Motonobu and Hennig, Philipp and Sejdinovic, Dino and Sriperumbudur, Bharath K},
  journal={arXiv preprint arXiv:1807.02582},
  year={2018}
}

@inproceedings{flaxman2005gradient,
author = {Flaxman, Abraham D. and Kalai, Adam Tauman and McMahan, H. Brendan},
title = {Online convex optimization in the bandit setting: gradient descent without a gradient},
year = {2005},
publisher = {Society for Industrial and Applied Mathematics},
booktitle = {ACM-SIAM Symposium on Discrete Algorithms},
}

@inproceedings{hazan2014bandit,
  title={Bandit convex optimization: Towards tight bounds},
  author={Hazan, Elad and Levy, Kfir},
  booktitle={Advances in Neural Information Processing Systems},
  year={2014}
}

@article{bubeck2021kernel,
  title={Kernel-based methods for bandit convex optimization},
  author={Bubeck, S{\'e}bastien and Eldan, Ronen and Lee, Yin Tat},
  journal={Journal of the ACM (JACM)},
  year={2021},
  publisher={ACM New York, NY}
}

@article{fokkema2024online,
  title={Online Newton method for bandit convex optimisation},
  author={Fokkema, Hidde and van der Hoeven, Dirk and Lattimore, Tor and Mayo, Jack J},
  journal={arXiv preprint arXiv:2406.06506},
  year={2024}
}

@article{hu2025optimal,
  title={Optimal computation budget allocation with Gaussian process regression},
  author={Hu, Mingjie and Xu, Jie and Chen, Chun-Hung and Hu, Jian-Qiang},
  journal={European Journal of Operational Research},
  year={2025},
}

@inproceedings{castiglioni2025bid,
author = {Castiglioni, Matteo and Nuara, Alessandro and Romano, Giulia and Spadaro, Giorgio and Trov\`{o}, Francesco and Gatti, Nicola},
title = {Safe Online Bid Optimization with Return on Investment and Budget Constraints},
year = {2025},
publisher = {Association for Computing Machinery},
booktitle = {ACM SIGKDD Conference on Knowledge Discovery and Data Mining},
}

@inproceedings{pmlr-v51-bogunovic16,
  title = 	 {Time-Varying Gaussian Process Bandit Optimization},
  author = 	 {Bogunovic, Ilija and Scarlett, Jonathan and Cevher, Volkan},
  booktitle = 	 {International Conference on Artificial Intelligence and Statistics},
  year = 	 {2016},
  editor = 	 {Gretton, Arthur and Robert, Christian C.},
  series = 	 {Proceedings of Machine Learning Research},
  publisher =    {PMLR},
}

@inproceedings{krause2021context,
 author = {Krause, Andreas and Ong, Cheng},
 booktitle = {Advances in Neural Information Processing Systems},
 title = {Contextual Gaussian Process Bandit Optimization},
 year = {2011}
}

@inproceedings{srinivas2010gaussian,
    title     = {Gaussian Process Optimization in the Bandit Setting:
                 No Regret and Experimental Design},
    author    = {Srinivas, Niranjan and Krause, Andreas and
                 Kakade, Sham M. and Seeger, Matthias},
    booktitle = {Proceedings of the 27th International Conference on
                 Machine Learning (ICML)},
    year      = {2010},
    pages     = {1015--1022}
  }

@inproceedings{chowdhury2017kernelized,
    title     = {On Kernelized Multi-armed Bandits},
    author    = {Chowdhury, Sayak Ray and Gopalan, Aditya},
    booktitle = {Proceedings of the 34th International Conference on
                 Machine Learning (ICML)},
    year      = {2017},
    pages     = {844--853}
  }

@inproceedings{krause2011contextual,
    title     = {Contextual {G}aussian Process Bandit Optimization},
    author    = {Krause, Andreas and Ong, Cheng Soon},
    booktitle = {Advances in Neural Information Processing Systems 24
                 (NeurIPS)},
    year      = {2011},
    pages     = {2447--2455}
  }

@book{matern1960spatial,
    title     = {Spatial Variation},
    author    = {Mat\'{e}rn, Bertil},
    year      = {1960},
    publisher = {Meddelanden fr\r{a}n Statens Skogsforskningsinstitut},
    note      = {Second edition: Springer Lecture Notes in Statistics, 1986}
  }

@InProceedings{pmlr-v247-suggala24a,
  title = 	 {Second Order Methods for Bandit Optimization and Control},
  author =       {Suggala, Arun and Sun, Y Jennifer and Netrapalli, Praneeth and Hazan, Elad},
  booktitle = 	 {Proceedings of Thirty Seventh Conference on Learning Theory},
  year = 	 {2024},
  series = 	 {Proceedings of Machine Learning Research},
  publisher =    {PMLR},
}

@article{auer2002finite,
    title   = {Finite-time Analysis of the Multiarmed Bandit Problem},
    author  = {Auer, Peter and Cesa-Bianchi, Nicol\`{o} and Fischer, Paul},
    journal = {Machine Learning},
    volume  = {47},
    number  = {2--3},
    pages   = {235--256},
    year    = {2002}
  }

@book{lattimore2020bandit,
    title     = {Bandit Algorithms},
    author    = {Lattimore, Tor and Szepesv\'{a}ri, Csaba},
    year      = {2020},
    publisher = {Cambridge University Press}
  }

@inproceedings{aggarwal2019autobidding,
    title     = {Autobidding with Constraints},
    author    = {Aggarwal, Gagan and Badanidiyuru, Ashwinkumar and
                 Mehta, Aranyak},
    booktitle = {Proceedings of the 15th International Conference on
                 Web and Internet Economics (WINE)},
    year      = {2019},
    pages     = {17--30}
  }

@inproceedings{deng2021towards,
    title     = {Towards Efficient Auctions in an Auto-bidding World},
    author    = {Deng, Yuan and Mao, Jieming and Mirrokni, Vahab and
                 Zuo, Song},
    booktitle = {Proceedings of the Web Conference (WWW)},
    year      = {2021},
    pages     = {3965--3973}
  }

@book{sutton1998reinforcement,
  title={Reinforcement learning: An introduction},
  author={Sutton, Richard S and Barto, Andrew G and others},
  volume={1},
  number={1},
  year={1998},
  publisher={MIT press Cambridge}
}

@article{naik2003understanding,
    title     = {Understanding the Impact of Synergy in Multimedia
                 Communications},
    author    = {Naik, Prasad A. and Raman, Kalyan},
    journal   = {Journal of Marketing Research},
    volume    = {40},
    number    = {4},
    pages     = {375--388},
    year      = {2003}
  }

@article{villar2015multi,
    title     = {Multi-armed Bandit Models for the Optimal Design of
                 Clinical Trials: Benefits and Challenges},
    author    = {Villar, Sof\'{i}a S. and Bowden, Jack and Wason, James},
    journal   = {Statistical Science},
    volume    = {30},
    number    = {2},
    pages     = {199--215},
    year      = {2015}
  }



\end{document}